\documentclass{article}

\usepackage{arxiv}

\usepackage[utf8]{inputenc} 
\usepackage[T1]{fontenc}    
\usepackage{hyperref}       
\usepackage{url}            
\usepackage{booktabs}       
\usepackage{amsfonts}       
\usepackage{nicefrac}       
\usepackage{microtype}      
\usepackage{lipsum}
\usepackage{graphicx}
\graphicspath{ {./images/} }
\usepackage{algorithm}
\usepackage{algpseudocode}
\usepackage{booktabs}   
\usepackage{amsmath}

\title{When Astronomy Meets AI: \texttt{Manazel} For Crescent Visibility Prediction in Morocco}

\author{
 Yassir Lairgi \\
  INSA Lyon\\
  LIRIS\\
  Villeurbanne, France \\
  \texttt{yassirlairgi@gmail.com} \\
}

\begin{document}
\maketitle
\begin{abstract}
The accurate determination of the beginning of each Hijri month is essential for religious, cultural, and administrative purposes. \texttt{Manazel}\footnote{The code and datasets are available at \url{https://github.com/lairgiyassir/manazel}} addresses this challenge in Morocco by leveraging 13 years of crescent visibility data to refine the ODEH criterion, a widely used standard for lunar crescent visibility prediction. The study integrates two key features, the Arc of Vision (ARCV) and the total width of the crescent (W), to enhance the accuracy of lunar visibility assessments. A machine learning approach utilizing the Logistic Regression algorithm is employed to classify crescent visibility conditions, achieving a predictive accuracy of 98.83\%. This data-driven methodology offers a robust and reliable framework for determining the start of the Hijri month, comparing different data classification tools, and improving the consistency of lunar calendar calculations in Morocco. The findings demonstrate the effectiveness of machine learning in astronomical applications and highlight the potential for further enhancements in the modeling of crescent visibility.
\end{abstract}


\section{Introduction}
The determination of the beginning of each Hijri month holds significant religious, cultural, and administrative importance in various Islamic societies \cite{nawawi2024hijri}. In Islamic countries, the visibility of the lunar crescent marks the start of each month in the Hijri calendar, which is essential for religious observances such as Ramadan and Eid celebrations. The criteria for crescent sighting vary between countries; some rely on direct human observation, while others use astronomical calculations or a combination of both \cite{sopa2022implementation}. For instance, Saudi Arabia often considers calculated visibility, while countries like Morocco adhere strictly to naked-eye observations. These differing approaches sometimes lead to discrepancies in the Hijri calendar between nations, highlighting the need for accurate prediction models.

Historically, methods for determining lunar crescent visibility date back to ancient Babylonian astronomy, where scholars developed empirical rules to forecast lunar phases \cite{article}. Over time, various visibility criteria have been introduced, with the ODEH criterion emerging as one of the most widely used and accurate models for predicting crescent visibility. Despite its effectiveness, the ODEH criterion has limitations when applied to Morocco’s specific conditions, where classes of visibility do not always provide definitive conclusions about whether the crescent will be seen or not.

\texttt{Manazel} seeks to enhance the precision of crescent visibility predictions by leveraging a comprehensive dataset spanning 13 years of lunar observations in Morocco. The significance of crescent visibility prediction is particularly high in Morocco, as the country relies exclusively on naked-eye observations to determine the start of each Hijri month. This reliance makes it crucial to refine predictive models to reduce uncertainties and improve accuracy.

To address these challenges, this study refines the ODEH criterion by incorporating two essential parameters: the Arc of Vision (ARCV) and the total width of the crescent (Wtot). These features provide additional insights into the complex dynamics of lunar sighting, aiding in more reliable assessments. Furthermore, this study introduces a machine learning-based approach to classify crescent visibility conditions. Utilizing the Logistic Regression algorithm, the research achieves an impressive predictive accuracy of 98.83\%, demonstrating the efficacy of data-driven methods in astronomical applications.

By tuning the ODEH model based on Morocco’s historical observations and applying machine learning techniques, this research offers a robust and reliable framework for determining the start of the Hijri month. The integration of AI-driven analysis enhances predictive reliability and opens new avenues for refining astronomical criteria through data science innovations. The findings underscore the relevance of advanced computational techniques in addressing long-standing challenges in lunar calendar determination, setting the stage for further enhancements in predictive modeling and methodological refinement.

\section{Related Works}
\subsection{Empirical Altitude–Azimuth Criteria}
The earliest methods rely on empirical relationships between the moon’s altitude and azimuth. Pioneered by Fotheringham \cite{fotheringham1910moon} and subsequently refined, Ilyas \cite{ilyas1987ancients}, and Fatoohi \cite{fatoohi1998first}, these criteria are typically expressed as polynomial equations that define a threshold for visibility. Although simple and based on direct observational data, these models suffer from being overly static; they often fail to capture the variability in atmospheric conditions, observer differences, and geographical diversity. For instance, Ilyas’s criterion has been criticized for underestimating the human eye’s capability, thereby limiting its practical predictive power. 

\subsection{Lunar Cycle Analysis}
Another approach focuses on the statistical examination of lunar cycle patterns \cite{rodzali2021relevansi} \cite{rahimi2019ketepatan}. These analyses explore the frequency of 29‑ versus 30‑day months and compare the resulting patterns with observational data. While they provide insight into the broader calendrical implications of crescent sightings, their main shortcoming lies in the inherent variability of lunar cycles. Each cycle is unique because of factors like sun–moon declination differences, observer location, and atmospheric influences; thus, the lunar cycle approach is less reliable for predicting individual crescent sightings or for forming a robust calendrical criterion. 

\subsection{Arc of Vision Versus Arc of Light (Elongation) Criteria}
Modern calendrical systems have also adopted criteria based on a combination of the arc of vision (the altitude difference between the moon and the sun) and the arc of light (or elongation) \cite{sopa2022implementation}. These methods aim to eliminate false-positive sightings by setting a lower boundary that positive observations must exceed. Yet, their rigidity—often designed specifically to support a calendrical framework—can lead to cases where real, but borderline, observations are disregarded. In particular, the criteria can be too conservative, especially when distinguishing between naked-eye and optical-aided observations

\subsection{Arc of Vision Versus Width Criteria}
Introduced by Bruin \cite{bruin1977first} and later adapted by researchers such as Yallop \cite{yallop1997method}, Odeh \cite{odeh2004new}, Qureshi \cite{qureshi2010new}, and Alrefay et al \cite{alrefay2018analysis}, these criteria extend the analysis by incorporating the lunar crescent’s width into the visibility equation. Although this method adds a valuable geometric dimension that can also predict observation windows, it is hampered by the difficulty in accurately predicting crescent visibility because the empirical conditions cannot reflect the actual crescent visibility in a country. 

\subsection{Lag Time–Based Criteria}
Lag time, the interval between sunset (or moonset) and the relevant lunar phase, has also been used as a predictor \cite{caldwell2012moonset}. While lag time is intuitively appealing (longer intervals are thought to correlate with easier visibility), its predictive power is undermined by low correlation coefficients and its sensitivity to observer latitude. In high-latitude regions, for example, the lunar crescent’s slanted path significantly alters lag time, rendering it less reliable as a sole or primary parameter in visibility criteria.

Despite the notable advancements achieved by state-of-the-art methods, these techniques are fundamentally limited by their reliance on static, one-size-fits-all thresholds. Such rigidity hinders their ability to accommodate the inherent variability of each country's data. Our work leverages a comprehensive 13-year dataset of crescent observations in Morocco and employs a data-driven machine learning framework to dynamically refine the ODEH criterion by adapting to the unique characteristics of Morocco 

\section{Proposed Method}
\label{sec:headings}
\subsection{Problem Statement}
Evaluating crescent visibility requires combining at least two parameters: one that reflects the intrinsic brightness of the crescent and another that accounts for its distance from the horizon, which is closely tied to atmospheric extinction \cite{odeh2004new}. Consequently, our supervised learning model incorporates the Arc of Vision (ARCV) and Crescent Width (W) as critical features. Odeh’s work \cite{odeh2004new} established an empirical equation using these parameters with specific thresholds to predict crescent visibility. In our study, \texttt{Manazel} automatically estimates a classifier for Morocco using these features, based on a dataset spanning 13 years of crescent observations.

\begin{figure}[h!]
\centering
\includegraphics[width=0.7\textwidth]{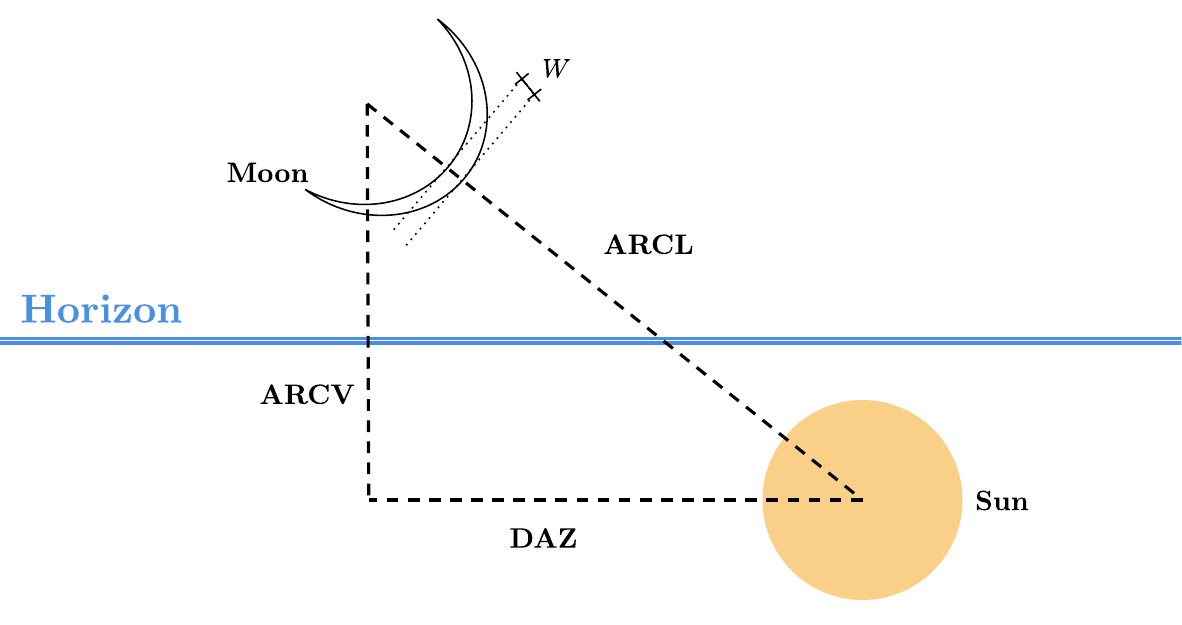}
\caption{\label{fig:crescent_vis_pred} Geometric parameters of the Sun and Moon during sunset.}
\end{figure}

Table \ref{tab:visibility_parameters_detailed} details the key parameters involved, such as ARCV and Crescent Width (W), providing clear descriptions of each. Moreover, Figure \ref{fig:crescent_vis_pred} visually demonstrates the geometric configuration of the Sun and Moon at sunset, which underpins the understanding of these parameters in the context of crescent visibility.

\begin{table}[ht]
\centering
\caption{Detailed Descriptions of Key Lunar Crescent Visibility Parameters}
\label{tab:visibility_parameters_detailed}
\begin{tabular}{lp{0.65\textwidth}}
\toprule
\textbf{Parameter} & \textbf{Detailed Description} \\
\midrule
Arc of Vision (ARCV) & Represents the vertical angular separation between the Sun and the Moon as seen by an observer. A higher ARCV typically means the Moon is positioned higher above the horizon, which minimizes atmospheric distortion and enhances the likelihood of detecting the crescent. \\
Relative Azimuth (DAZ) & Denotes the horizontal angular difference between the Sun and the Moon along the observer's horizon. This measurement reflects how the Moon is positioned relative to the Sun in the sky, affecting the contrast and orientation of the illuminated crescent under various observational conditions. \\
Crescent Width (W) & Measures the extent of the illuminated (lit) portion of the Moon along its diameter. A larger width indicates a broader and potentially brighter crescent, making it easier to observe, whereas a narrower width suggests a thinner, fainter crescent that may be more challenging to detect. \\
\bottomrule
\end{tabular}
\end{table}

\subsection{Data Preparation}
The dataset was constructed by first exploring the official website of the Moroccan Ministry of Awqaf\footnote{\url{https://www.habous.gov.ma}} to track and record the start of each Hijri month. For each confirmed sighting of the crescent, the astronomical parameters ARCV and W of the preceding day were computed and labeled as 1 (seen). In cases where the crescent was not observed on the 29th night, indicative of a 30-day month, the corresponding astronomical parameters were recorded two days earlier and labeled as 0 (not seen). This procedure yielded a balanced dataset of 257 observations, comprising 153 instances where the Hilal was seen and 104 instances where it was not. Table~\ref{tab:dataset_summary} presents the summary statistics for the key variables in the dataset.

\begin{table}[h!]
\centering
\caption{Summary Statistics of the Moroccan Crescent Visibility Dataset}
{%
\begin{tabular}{lrrrrrrrr}
\hline
\textbf{Variable} & \textbf{Count} & \textbf{Mean} & \textbf{Std} & \textbf{Min} & \textbf{25th} & \textbf{50th} & \textbf{75th} & \textbf{Max} \\
\hline
arcv    & 257 & 9.5620  & 5.0969  & 0.6054 & 5.1224 & 9.0396 & 13.1915 & 22.5345 \\
\(W\) & 257 & 0.4911  & 0.3949  & 0.0057 & 0.1557 & 0.4037 & 0.7661  & 1.6906 \\
output  & 257 & 0.5953  & 0.4918  & 0      & 0      & 1      & 1       & 1 \\
\hline
\end{tabular}%
}
\label{tab:dataset_summary}
\end{table}

\subsection{Exploratory Analysis of Moroccan Crescent Visibility Data}
\begin{figure}[h!]
\centering
\includegraphics[width=0.7\textwidth]{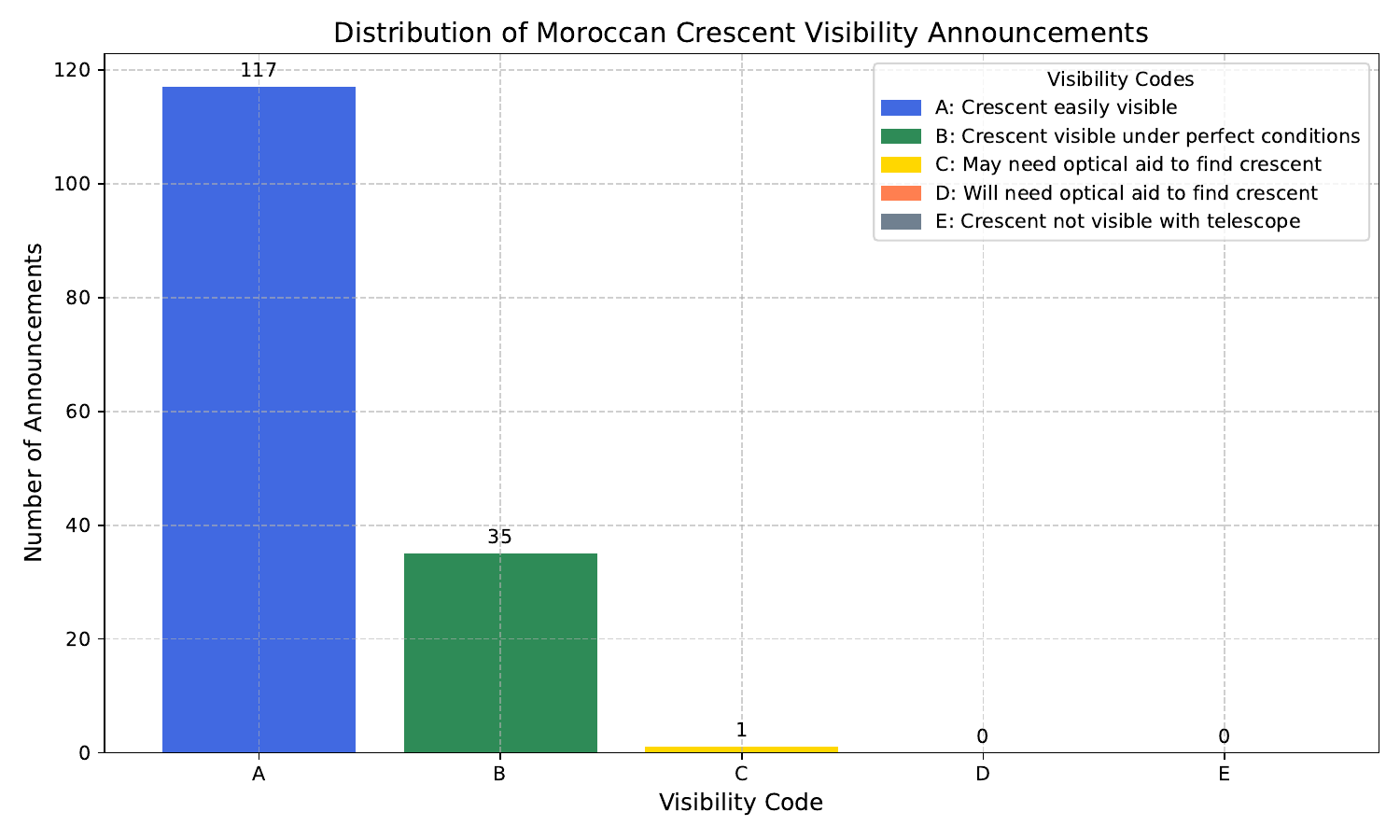}
\caption{\label{fig:crescent_vis_dist} Distribution of Moroccan Crescent Visibility Announcements.}
\end{figure}

According to the ODEH criterion, crescent sightings in Morocco have been assessed as remarkably successful over the past 13 years (cf. Figure \ref{fig:crescent_vis_dist}). Most announcements have been classified as A and B, which is consistent with using naked-eye observations to determine the Hijri calendar in Morocco. However, a notable limitation has been identified: when the ODEH criterion classifies crescent visibility as B, it becomes difficult to predict whether the crescent will be visible definitively. In the corresponding plot, ambiguity is observed in the B classifications, as a clear decision regarding visibility cannot be reached. It is further demonstrated that the A classifications exhibit the highest probability of confirmed visibility, followed by the B classifications. Motivated by these observations, a deeper analysis of Moroccan crescent visibility over the past 13 years was conducted, and a machine learning model was subsequently trained to refine the ODEH criterion based on the ARCV and W features.

\subsection{Model Training}
Five classification models: Logistic Regression, Decision Tree, Random Forest, Support Vector Machine, and K-Nearest Neighbors were configured with specific hyperparameter grids, and hyperparameter tuning was conducted with 4-fold cross-validation to systematically evaluate parameter combinations and select the optimal configuration based on accuracy. The best estimator for each model was subsequently used to generate cross-validated predictions, from which overall accuracy and detailed classification reports were computed (cf. Table \ref{tab:hyperparameter_tuning}).

Although all models exhibited comparable performance, Logistic Regression is the preferred model thanks to its inherent interpretability and slightly superior overall accuracy. As summarized in Table \ref{tab:hyperparameter_tuning}, Logistic Regression achieved a best CV score of 0.9922 and an overall CV accuracy of 0.9883, which is marginally higher than those of the other evaluated models.

\begin{table}[ht]
\centering
\caption{Hyperparameter Tuning Results using Cross-Validation}
\resizebox{\linewidth}{!}{%
\begin{tabular}{lccc}
\hline
\textbf{Model} & \textbf{Best Hyperparameters} & \textbf{Best CV Score} & \textbf{Overall CV Accuracy} \\
\hline
Logistic Regression & \{C: 100, solver: lbfgs\} & \textbf{0.9922} & \textbf{0.9883} \\
Decision Tree       & \{max\_depth: 5, min\_samples\_split: 2\} & 0.9844 & 0.9767 \\
Random Forest       & \{max\_depth: 3, min\_samples\_split: 10, n\_estimators: 50\} & 0.9883 & 0.9844 \\
SVM                 & \{C: 100, kernel: linear\} & 0.9844 & 0.9844 \\
KNN                 & \{n\_neighbors: 3, weights: distance\} & \textbf{0.9922} & \textbf{0.9844} \\
\hline
\end{tabular}%
}
\label{tab:hyperparameter_tuning}
\end{table}

\subsection{Crescent Visibility For the Determination of Hijri Month Start Date in Morocco}

Algorithm \ref{alg:hilal_visibility} presents a procedure for determining the start date of a Hijri month based on crescent visibility observations. In this approach, the initial Hijri date (with the day set to 1) is first converted to a Gregorian baseline, assuming a default month length of 29 days. An iterative process is then employed, during which the Gregorian date is adjusted and astronomical parameters such as ARCV and W are computed for Rabat. These parameters are subsequently input into the selected model that assesses the likelihood of the crescent visibility. When the model indicates a positive detection, the iteration is halted and the corresponding Gregorian date is designated as the first day of the Hijri month.

\begin{algorithm}[ht]
\caption{Determination of Hijri Month Start Date via Hilal Visibility}
\label{alg:hilal_visibility}
\begin{algorithmic}[1] 
\State \textbf{Input:} Hijri date $(H_{year}, H_{month}, H_{day}=1)$, location (Rabat), maximum iterations $N_{\max}$
\State \textbf{Output:} Gregorian date corresponding to the first day of the Hijri month
\medskip
\State \textbf{Step 1: Initial Gregorian Conversion}
\State Convert the Hijri date $(H_{year}, H_{month}, 1)$ to a Gregorian date $G_{\text{base}}$, assuming a default Hijri month duration of 29 days.
\medskip
\State \textbf{Step 2: Iterative Hilal Visibility Check}
\State Initialize offset $\Delta \gets -1$.
\While{$\Delta \leq N_{\max}$}
    \State Calculate the doubt night: $G_{\text{doubt}} \gets G_{\text{base}} + \Delta$.
    \State Compute the astronomical parameters (e.g., ARCV, $W$) for the location (Rabat).
    \State Evaluate the predictive model $f(\text{ARCV}, W)$.
    \If{$f(\text{ARCV}, W) = 1$}
         \State \textbf{break} the loop.
    \EndIf
    \State Increment the offset: $\Delta \gets \Delta + 1$.
\EndWhile
\medskip
\State \textbf{Step 3: Determining the First Day of the Month}
\State The first day of the Hijri month is set as $G_{\text{first}} \gets G_{\text{doubt}} + 1$.
\State \Return $G_{\text{first}}$.
\end{algorithmic}
\end{algorithm}

\section{Conclusion}
Manazel has demonstrated that integrating astronomical observations with machine learning techniques can significantly enhance the accuracy and reliability of crescent visibility predictions for determining the start of the Hijri month in Morocco. 

A comprehensive dataset spanning 13 years was constructed by systematically recording Hilal sightings from the official Moroccan Ministry of Awqaf and by computing key parameters such as the Arc of Vision (ARCV) and the crescent width (W). Several classification models were evaluated using cross-validation, and despite comparable performance among the candidates, Logistic Regression was ultimately selected due to its superior predictive accuracy (overall CV accuracy of 98.83\%) and inherent interpretability. 

Furthermore, an algorithm was developed to convert the Hijri date to its corresponding Gregorian date based on iterative visibility checks, thereby refining the traditional ODEH criterion. Future work will focus on incorporating additional atmospheric and observational variables to further enhance the model’s performance and robustness.

\section*{Acknowledgements}

The author gratefully acknowledge that the dataset utilized in this study was meticulously prepared and curated by Chaymae Majdoubi. Her dedicated efforts in collecting and refining 13 years of crescent visibility observations were instrumental in enabling the development and rigorous evaluation of the \texttt{Manazel} model.

\bibliographystyle{unsrt}  
\bibliography{references}  

\end{document}